\documentclass[10pt,twocolumn,letterpaper]{article}

\usepackage{subfig}

\usepackage{iccv}
\usepackage{times}
\usepackage{epsfig}
\usepackage{graphicx}
\usepackage{amsmath}
\usepackage{amssymb}

\usepackage{booktabs}

\usepackage{multirow}
\usepackage{multicol}

\newcommand\blfootnote[1]{%
  \begingroup
  \renewcommand\thefootnote{}\footnote{#1}%
  \addtocounter{footnote}{-1}%
  \endgroup
}

\usepackage[pagebackref=true,breaklinks=true,letterpaper=true,colorlinks,bookmarks=false]{hyperref}

\iccvfinalcopy % *** Uncomment this line for the final submission

\def\iccvPaperID{****} % *** Enter the ICCV Paper ID here

\begin{document}

%%%%%%%%% TITLE
\title{Unconditional Scene Graph Generation}

\author{Sarthak Garg $^{1,}$\thanks{The first two authors contributed equally to this work} \hspace{1.1cm}
%\and
Helisa Dhamo $^{1,}$\footnotemark[1]  \hspace{0.9cm}
%\and
Azade Farshad $^{1}$  \\ \vspace{-0.25cm} \\ %\hspace{1.8cm}
%\and 
\hspace{0.1cm}
Sabrina Musatian $^{1}$    \hspace{0.9cm}
%\and
Nassir Navab $^{1}$ \hspace{0.9cm}
%\and
Federico Tombari $^{1,2}$
\and
$^{1}$ Technische Universit\"at M\"unchen
\hspace{0.25cm}
$^{2}$ Google
}

\maketitle
%\ificcvfinal\thispagestyle{empty}\fi

\newcommand{\nameMethod}{SceneGraphGen }

%%%%%%%%% ABSTRACT
\begin{abstract}
 Despite recent advancements in single-domain or single-object image generation, it is still challenging to generate complex scenes containing diverse, multiple objects and their interactions. Scene graphs, composed of nodes as objects and directed-edges as relationships among objects, offer an alternative representation of a scene that is more semantically grounded than images. We hypothesize that a generative model for scene graphs might be able to learn the underlying semantic structure of real-world scenes more effectively than images, and hence, generate realistic novel scenes in the form of scene graphs. In this work, we explore a new task for the unconditional generation of semantic scene graphs. We develop a deep auto-regressive model called \nameMethod which can directly learn the probability distribution over labelled and directed graphs using a hierarchical recurrent architecture. The model takes a seed object as input and generates a scene graph in a sequence of steps, each step generating an object node, followed by a sequence of relationship edges connecting to the previous nodes. We show that the scene graphs generated by \nameMethod are diverse and follow the semantic patterns of real-world scenes. Additionally, we demonstrate the application of the generated graphs in image synthesis, anomaly detection and scene graph completion. \blfootnote{\url{https://SceneGraphGen.github.io/}}
\end{abstract}

\section{Introduction}

Scene graphs encompass a representation that describes a scene, where nodes are categorical object instances and the edges describe categorical relationships between them. This representation allows for an extended high-level understanding of the scenes, which goes beyond object-level reasoning. The computer vision community has explored various approaches for scene graph generation from images \cite{xu2017scenegraph,newell2017pixels} as well as tasks in which this representation has proven to be suitable, such as image retrieval \cite{johnson15} and VQA \cite{ghosh2019generating}. A scene graph allows for modular, specified, and high-level semantic control on the image components, which makes it a good representation for semantic-driven image generation \cite{johnson2018image} and manipulation \cite{Dhamo2020cvpr}. 

\begin{figure}[t]
\begin{center}
   \includegraphics[width=\linewidth]{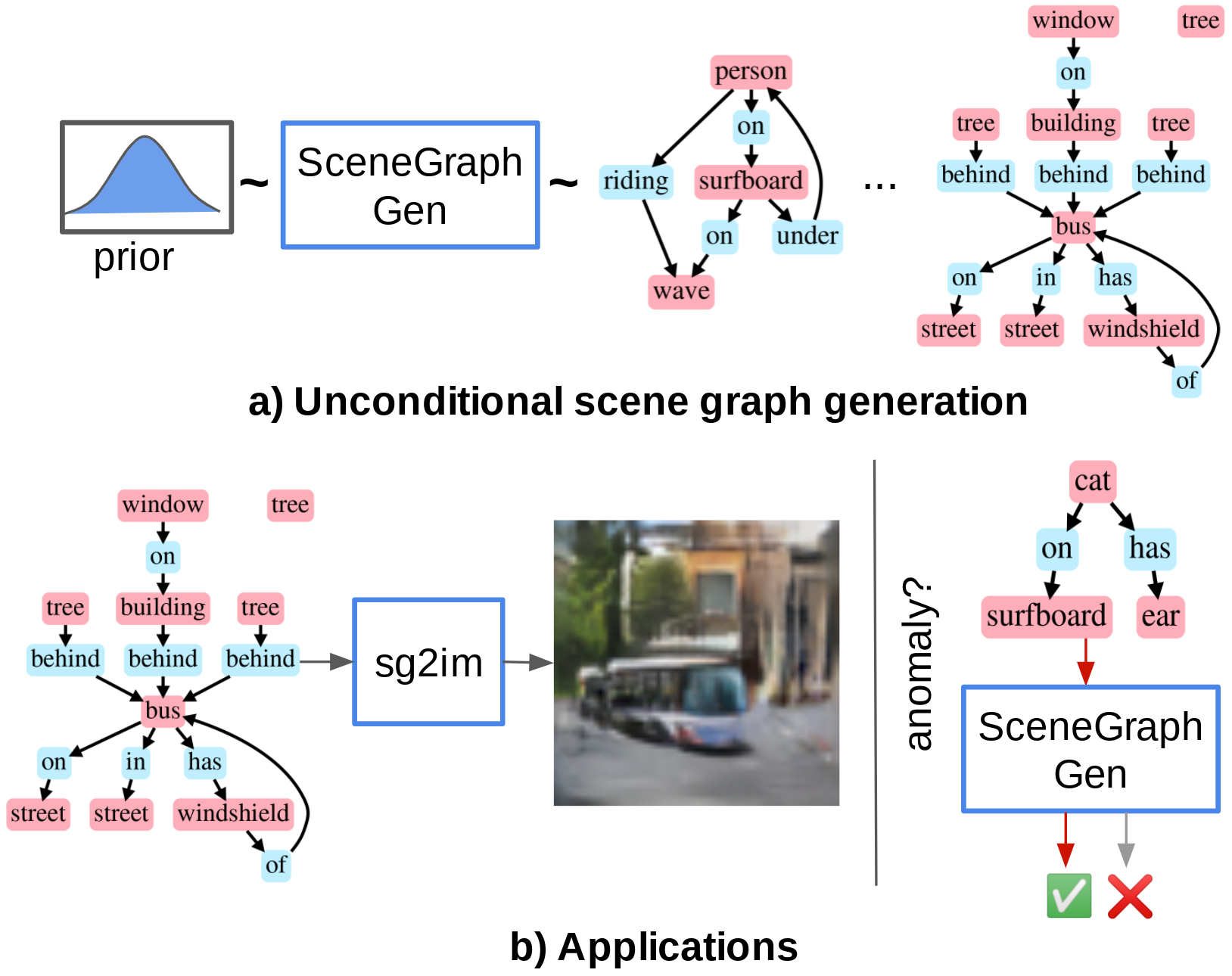}
\end{center}
   \caption{\textbf{Method overview.} \textbf{a)} \nameMethod generates scene graphs unconditionally from a randomly sampled seed object. \textbf{b)} Applications in \emph{Left:} translation of a generated scene graph to an image, using an off-the-shelf graph-to-image network \emph{Right:} detection of out-of-distribution samples}
\label{fig:teaser}
\end{figure}

A less explored field has been \emph{unconditional generation of scene graphs}, \ie generating a scene graph without any input image, but instead, from a \textit{random} input. Such scene graph modelling can aid learning patterns from real scenes, such as object co-occurrence, relative placements and interactions.  
In this paper, scene graph generation is explored under the light of generative models, with the aim to generate novel and realistic instances of scene graphs \emph{unconditionally}. Recent work explored generation of relational graphs~\cite{Wang2019PlanITPA} or probabilistic grammar~\cite{kar2019metasim}, designed for a particular domain. To the best of our knowledge, we are the first to investigate the use of a generative model for generating semantic, language-based scene graphs~\cite{johnson15,krishna2017visual}. As scene graphs describe scenes, it is possible to translate the generated graph to another domain, using state-of-the-art models specialized, for instance, in the graph-to-image task~\cite{johnson2018image}. In the context of unconditional image generation, recent works demonstrate impressive results mostly in object-centric images that contain one main subject, or uni-modal distributions, such as datasets of faces or cars. On the other hand, complex and diverse scenes which contain multiple objects are more difficult to capture by these models. We show in our experiments that modelling unconditional image scene generation through scene graphs instead leads to more distinguishable object instances, as it enables understanding of complex and often abstract semantic concepts such as objects, their interactions, and attributes. Additionally, such a generative model can detect out-of-distribution scene graphs and complete partial scene graphs.

Recently, deep generative models have been proposed for graph data \cite{you2018graphrnn,grover2019graphite,lggan,simonovsky2018graphvae,netgan}, which aim to synthesize realistic graphs of a certain domain while capturing graph patterns, such as degree distribution and clustering. However, each model comes with caveats making it unsuitable for certain applications. The size of scene graphs often varies significantly, the object and relationship categories are inherently unbalanced, and the edges are directed. For this purpose, we develop a specialized model called \nameMethod (Figure \ref{fig:teaser}). The overall auto-regressive structure is inspired from GraphRNN \cite{you2018graphrnn}, as it accommodates varying graph sizes, unlike \cite{lggan, simonovsky2018graphvae, molgan}. Specifically, the model is adapted to consume categories for nodes and edges, as well as to support the direction of the edges. In this auto-regressive formulation, the scene graphs are represented as a sequence of sequences. The history of the sequence is carried in a hidden state using a Gated Recurrent Unit (GRU) \cite{cho2014properties}, which is used to generate categorical distribution over the nodes and edges at each step, from which the node and edge categories can be sampled. The nodes are generated using a multi-layer perceptron (MLP) and the edges are generated sequentially using a GRU.

Since unconditional scene graph generation is a new task, metrics to evaluate the quality of the generated graphs have not been proposed yet. Thus, following \cite{you2018graphrnn} we leverage a Maximum Mean Discrepancy (MMD) metric, adapted with a random-walk graph kernel and a node kernel, which are appropriate for the scene graph structure. We verify the validity of these kernels using sets of corrupted datasets. 

Our contributions can be summarized as follows:
\begin{itemize}
    \item We introduce \nameMethod to tackle the unexplored task of unconditional semantic scene graph generation. We adopt a graph auto-regressive model to enable processing the scene graphs structure.
    \item We demonstrate the use of the learned scene graph model in three applications, namely image generation, anomaly detection, and scene graph completion.
    \item We propose and verify an MMD metric to evaluate the generated scene graphs, which operates on the node and graph level.
\end{itemize}

We evaluate our model on Visual Genome \cite{krishna2017visual} and show that it can generate semantically plausible scene graphs. We show how these scene graphs can transfer to novel images, using state-of-the-art scene graph to image models~\cite{johnson2018image}. Additionally, we show that the model can be used to detect unusual scene graphs and extend incomplete scene graphs. 
\section{Related Work}

 We discuss works related to scene graph inference, their applications, and generative models on graphs in general. 
 
\paragraph{Scene graphs and applications}

 Scene graphs as proposed in~\cite{johnson15} offer a semantic description for an image, in the form of semantic labels for objects and their relationships. Large-scale datasets, \eg Visual Genome \cite{krishna2017visual} annotated with scene graphs enabled deep learning tasks. A line of works explores scene graph prediction conditioned on an image \cite{xu2017scenegraph,newell2017pixels,yang2018graph,li2018factorizable,zellers2018neural} or point clouds~\cite{3DSSG2020,Wu_2021_CVPR}. Most works first detect the objects in the scene, and then reason about their relationships. In contrast, our work focuses on unconditional graph generation-- \ie does not rely on an input at test time -- to model the scene graph distribution.

\begin{figure*}
\begin{minipage}{0.6\textwidth}
    \includegraphics[width=\linewidth]{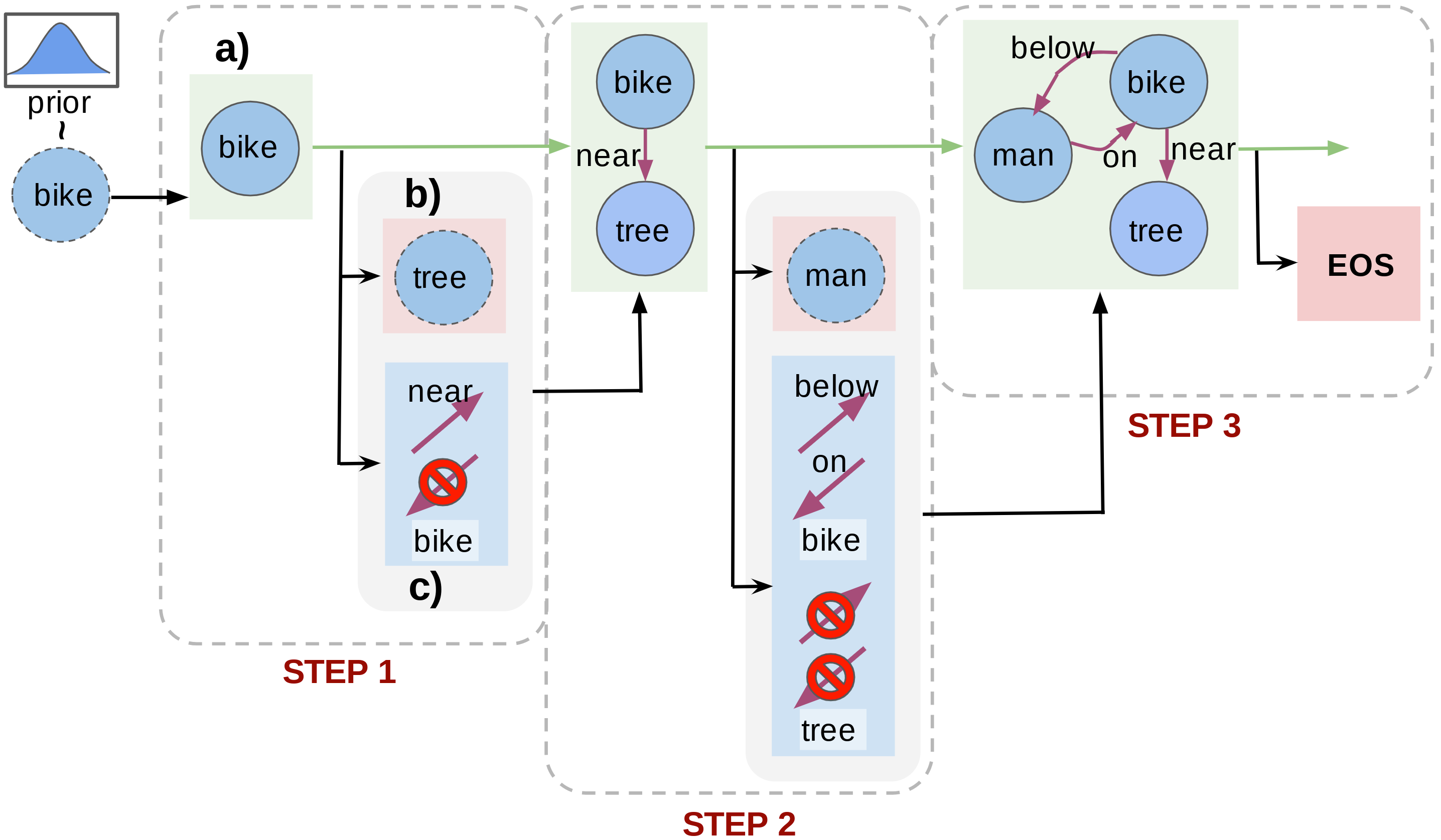}
    \end{minipage}
    \begin{minipage}{0.4\textwidth}
    \includegraphics[width=\linewidth]{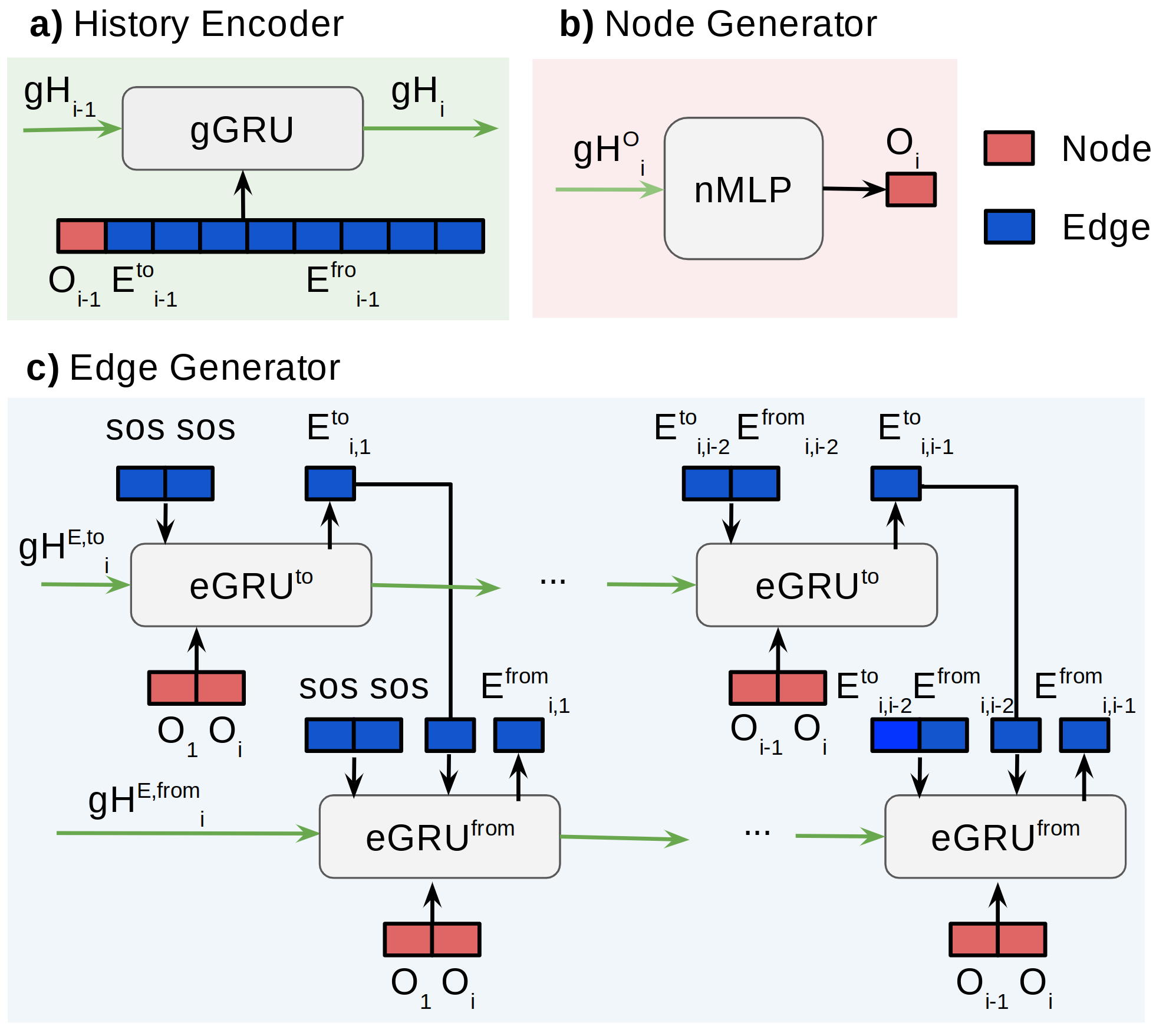}
    \end{minipage}
    \caption{\textbf{Illustration of the SceneGraphGen architecture.} \emph{Left:} The overview of the auto-regressive generation process, which takes the current graph sequence in each step and generates a new node and a set of connecting edges. \emph{Right:} Architecture details of each of the modules: \textbf{a)} history encoder, \textbf{b)} node generator, \textbf{c)} edge generator.}
    \label{fig:sgrnn}
\end{figure*}

Scene graphs can be used for a variety of tasks. Johnson \etal~\cite{johnson2018image} propose image generation from a scene graph, further explored in an interactive setup for generation \cite{ashual2019specifying} and semantic manipulation \cite{Dhamo2020cvpr}. Wang \etal~\cite{Wang2019PlanITPA} explore relational graphs for indoor scene planing. Other works exploit scene graphs for image and domain-agnostic retrieval \cite{johnson15,3DSSG2020}. Often scene graphs are combined with language, such as in Visual Question Answering (VQA)~\cite{ghosh2019generating,yang2019scene} or prediction of object type, given query locations~\cite{zhou2019scenegraphnet}.

\paragraph{Generative models on graphs} 
Traditional approaches~\cite{erdos59a,ergm,kronecker,sbm} are engineered to capture some specific patterns, often domain specific, and they fall short in generalizing to all graph patterns. 

Deep auto-regressive models for graph generation \cite{li2018learning,liao2019gran,you2018graphrnn,ingraham2019generative} are typically flexible in number of nodes, but need to impose a node ordering. 
GraphRNN \cite{you2018graphrnn} represents the graphs as a sequence of sequences, and uses a hierarchical GRU architecture to model the node and edge dependencies. The model is scalable with $O(N^2)$ complexity and can output variable size graphs. They utilize a breadth first search (BFS) ordering to significantly reduce the possible orderings. However, the paper only addresses generation of unlabelled graphs.

Another line of work uses Variational Autoencoders (VAE), which embed the graphs into a vector~\cite{simonovsky2018graphvae}, junction tree of node clusters~\cite{jin2019junction} or combine VAE with autoregressive approach \cite{liu2018constrained}. These models usually enable graph modeling with attributes/categories, but the model captures inexact likelihood and do not scale well to larger graphs. 

GAN based works are limited to either a single sample \cite{netgan}, or a small and fixed size graph \cite{lggan,molgan}, making them unsuitable for variable-sized and diverse scene graphs. Additionally, they struggle with training stability.

In this work, we explore an auto-regressive approach which allows for a flexible number of graph nodes. Unlike GraphRNN, our method supports semantic labels for nodes and edges, models directed edges, and leverages an edge generation GRU which is aware of the node categories. 

Some recent works explore generative tasks with similar scene structures (relational graphs \cite{Wang2019PlanITPA}, probabilistic grammar \cite{kar2019metasim,devaranjan2020metasim2}). Our focus is however on semantic graphs associated with in-the-wild images, which are considerably more diverse in node/edge labels and exhibit less regular graph size and connectivity patterns, compared to the synthetic datasets utilized in other works.

\section{Methodology}

Given a set of $n$ scene graph samples $\mathbb{G}_s = \{G_s^1, G_s^2, .., G_s^n\}$, which are assumed to represent the data distribution of scene graphs $p_{data}(G_s)$, the goal is to learn a generative model from dataset $\mathbb{G}_s$, which can generate new samples of scene graphs. A scene graph sample $G_s$ corresponding to an image $I$ is defined as $G_s = (\mathcal{O}, \mathcal{E})$ by the set of objects (nodes) $\mathcal{O}$ and relationships (edges) $\mathcal{E}$. Each object $o_i$ from the set $\mathcal{O} = \{o_1, o_2, .., o_m\}$ holds an object category  $o_i \in \mathcal{C}$, where $\mathcal{C} = \{1, 2, .., C\}$. The edges are an ordered triplet $\mathcal{E} \subseteq \{(o_i,r_k, o_j)|o_i, o_j \in \mathcal{O}, o_i\neq o_j\}$ denoting the directed edge from $o_i$ to $o_j$, where $r_k \in \mathcal{R}$ is the relationship category between the objects, $\mathcal{R}=\{1, 2, .., R\}$ (\eg \texttt{man} - \texttt{wearing} - \texttt{shirt}). We explore this task via an auto-regressive model, which enables flexibility in number of nodes and graph connectivity. We give a network overview in section \ref{sec:overview}, then describe each component in more detail in the proceeding sections.

\subsection{Model formulation}
\label{sec:overview}

\nameMethod learns a distribution $p_{\phi}(G_s)$ over scene graphs which is close to $p_{data}(G_s)$. For an auto-regressive formulation, we convert scene graphs to a sequence representation. Under a permutation $\pi$, the node set $\mathcal{O}$ becomes a sequence $O=(\pi(o_1), \pi(o_2), .., \pi(o_m))$. The edges $\mathcal{E}$ are represented using two upper triangular sparse matrices $E^{from}$ and $E^{to}$, with relationship label $r_k$ at index $(i, j)$ if $(\pi(o_i), r_k, \pi(o_j)) \in \mathcal{E}$ and $(\pi(o_j), r_k, \pi(o_i)) \in \mathcal{E}$ respectively. The matrices correspond to the two possible edges (to and from) between any node pair. A scene graph $G_s$ can, thus, be represented as a sequence $X = (O, E^{to}, E^{from})$.  Each element of the sequence $X$ is itself a sequence, given as $X_i = (O_i, E^{to}_i, E^{from}_i)$, where $O_i$ is the object node, $E^{to}_i$ and $E^{from}_i$ are the sequence of edges between $O_i$ and previous nodes. We translate the task of learning $p_{\phi}(G_s)$ to learning a sequence distribution $p_{\phi}(X)$, that can be modeled auto-regressively. The probability over sequence $X$ is decomposed into successive conditionals
\begin{equation}
\label{equation:sggen1}
    p_{\phi}(X) = p(X_1)\prod_{i=2}^{n}
    p_{\phi}(X_i|X_{<i}).
\end{equation}
$X_{<i} = (X_1, .., X_{i-1})$ depicts the partial scene graph sequence up to step $i$. We further split each conditional $p_{\phi}(X_i|X_{<i})$ into three parts for each of the components as
\begin{equation}
\label{equation:sggen2}
\begin{split}
    p_{\phi}(X) = p(O_1)\prod_{i=2}^{n}
    & p_{\phi_1}(O_i|X_{<i})p_{\phi_2}(E^{to}_i|O_i, X_{<i})\\
    & p_{\phi_3}(E^{from}_i|O_i, E^{to}_i, X_{<i}).
\end{split}
\end{equation}

We model Equation \ref{equation:sggen2} and learn the complete probability distribution over scene graph sequences $p_{\phi}(X)$. We do not make any conditional independence assumptions in our model formulation, which allows SceneGraphGen to potentially capture all the complex object and relationship dependencies present in the data. $p(O_1)$ is an assumed prior distribution over the first node, \eg the categorical distribution over the object occurrences. The auto-regressive modeling is performed in three stages.
\begin{enumerate}
    \item History Encoding: the information from the previous steps $X_{<i}$ is captured in a hidden representation $h_i$. This stage models $h_i = \text{Encoder}(X_{<i})$.
    \item Node generation: the hidden state is used to generate the next object node. This stage models $p_{\phi_1}(O_i|h_i)$.
    \item Edge generation: the hidden state, along with the explicit node information is used to sequentially generate the edges. This stage models $p_{\phi_2}(E^{to}_i|O_i, h_i)$ and $p_{\phi_3}(E^{from}_i|O_i, E^{to}_i, h_i)$.
\end{enumerate}
The overall architecture and the each of the components are schematically depicted in Figure \ref{fig:sgrnn}. Now, we describe each of the stages in detail.

\subsection{History encoding}

To carry the past information $X_{<i}$ at each time step $i$, we use a Gated Recurrent Unit (GRU) architecture. We harness three separate GRUs, which are specialized to carry information for each of the three outputs $O_i$, $E^{from}_i$ and $E^{to}_i$. This allows for decoupling of the information for the three outputs. Note that the parameters of each of the three GRUs are different but shared across all time steps. We call these networks $\text{gGRU}_O$, $\text{gGRU}_{E^{from}}$ and $\text{gGRU}_{E^{to}}$. The initial hidden states of the GRUs are taken to be a zero (empty) vector. In further steps, the input to all the three GRUs is the past sequence $X_{i-1}$, which is taken as the concatenation of the three outputs from the previous step $i-1$, \ie $O_{i-1}$, $E^{from}_{i-1}$ and $E^{to}_{i-1}$. For the first step, the edge inputs are empty (zero), and the node is sampled from a prior distribution. This prior is computed based on the occurrence of first node category for the selected ordering strategy.

\subsection{Node generation}

We use a multi-layer perceptron (MLP) which takes the hidden state from $\text{gGRU}_O$ as input, and outputs a categorical distribution over the object categories $\mathcal{C}$, as prediction scores $\theta^O_i$. We call this network nMLP. The parameters of nMLP are shared across all time steps. The node $O_i$ is then sampled from this distribution. The generation of new nodes (and hence, the scene graph) is halted when an EOS (end of sequence) token is encountered.

\subsection{Node-aware edge generation}

The output at each step $i$ is itself a \textit{sequence} of edges $E^{from}_i$. This forms a sequence of sequences, which in turn needs a sequential model for each step of the sequence.  For this hierarchical requirement, we use a GRU which shares the parameters, not only for all time steps $i$, but also for each step inside that sequence $i$, which we call $j$. A similar GRU is used to produce $E^{to}_i$ at each step $i$. We call these networks $\text{eGRU}_{E^{from}}$ and $\text{eGRU}_{E^{to}}$. 
The initial hidden state for these GRUs is taken as the hidden state resulting from $\text{gGRU}_{E^{from}}$ and $\text{gGRU}_{E^{to}}$ networks respectively. The initial input to these GRUs is an SOS (start of sequence) token, which starts the generation process. 
For any step $j$, the input for $\text{eGRU}_{E^{to}}$ is the concatenation of four elements, the edges from the previous step $E^{from}_{i, j-1}$ and $E^{to}_{i, j-1}$, the node $O_i$ at current step $i$, as well as the node $O_j$ from the $j^{th}$ time step. 
The additional information about the nodes provides \textit{additional conditioning} for the edge generation decision. This design choice is inspired from the high predictability of co-occurrence of relationships for a given object pair, \ie additional knowledge of node pairs might facilitate relationship predictions.
From this input, the GRUs generate categorical distribution parameters over the relation categories $\mathcal{R}$ (and a no-edge category), given by $\theta^{E^{to}}_{i, j}$. The next edge $E^{to}_{i, j}$ is then sampled from this distribution.
The inputs to the $\text{eGRU}_{E^{from}}$ at step $j$ are the same inputs as for $\text{eGRU}_{E^{to}}$, together with the edge $E^{to}_{i, j}$ generated by  $\text{eGRU}_{E^{to}}$ at the current step $j$. The next edge $E^{from}_{i, j}$ is generated by $\text{eGRU}_{E^{from}}$ in a similar fashion.
This process is carried out for $i-1$ steps, which is the number of previous nodes that can connect to the current node $O_i$. 

\subsection{Loss objective}
We learn the parameters $\phi$ of the model $p_{\phi}(X)$ by maximizing the likelihood of training data $\mathbb{G}$ on the model. The training is performed by using teacher forcing, \ie at each step $i$, we use the ground truth sequences $O_i, E^{to}_i, E^{from}_i$ instead of sampling from the model's prediction scores $\theta^{O}_{i}, \theta^{E_{from}}_{i}, \theta^{E_{to}}_{i}$. The loss is computed using cross-entropy (CE) between the predicted scores and the ground truth sequence at each step, and then summing it up. For a sample scene graph sequence $X$, the loss is given by
\begin{equation}
\mathcal{L}(X; \phi) = \mathcal{L}_{O}(O; \phi_1) + \mathcal{L}_{E}(E^{to}_i, E^{from}_i; \phi_2, \phi_3)
\end{equation}
\begin{equation}
\mathcal{L}_{O} = \sum_{i=2}^{n} CE(\theta^O_i, O_i)
\end{equation}
\begin{equation}
\mathcal{L}_{E} = \sum_{i=2}^{n}\sum_{j=1}^{i-1}CE(\theta^{E_{to}}_{i, j},  E^{to}_{i,j})
    +\sum_{i=2}^{n}\sum_{j=1}^{i-1}CE(\theta^{E_{from}}_{i, j},  E^{from}_{i,j})
\end{equation}

\subsection{Inference}

After learning the distribution over scene graph sequences $p_{\phi}(X)$, SceneGraphGen can be used to sample new scene graph instances. The sampling procedure essentially follows the auto-regressive generation formulation described so far. First, we sample the first object node from the prior distribution which makes the first sequence $X_1$. The prior distribution is empirically computed from the training set, based on the occurrences of first node when ordering the nodes using the ordering scheme chosen during training. In other steps we use the previous sequence $X_{i-1}$ as input to compute the hidden states using the three gGRU's. These hidden states are used to generate the next node $O_i$ using nMLP by sampling from the categorical output. Similarly, we generate the sequences of edges $E^{to}_i$ and  $E^{from}_i$ using $\text{eGRU}_{E^{to}}$ and $\text{eGRU}_{E^{from}}$ respectively from the respective categorical outputs. The node and sequence of edges are combined to form the next sequence $X_i$. This process is continued until the node generator outputs an EOS token.

SceneGraphGen can additionally be used to evaluate the likelihood of a given sample. For a given scene graph converted to a sequence $X$, each element of the sequence $X_{i-1}$ is passed to the network to obtain the node categorical outputs $\theta^O_i$ as well as edge categorical output sequences $\theta^{E_{from}}_{i}$ and $\theta^{E_{to}}_{i}$. Then, we pick the probability of the true node/edge category for each output and take the negative of the sum of logarithm over the whole sequence $X$ to obtain the negative log-likelihood (NLL). This NLL value depicts the 'unlikeliness' of occurrence of a given scene graph sample. We use the NLL to detect anomalies in Section \ref{section:anomaly}.

\subsection{Implementation details}

\nameMethod is trained for 300 epochs, with a batch size of 256, where 256 batches are sampled with replacement per epoch. The initial learning rate is 0.001. We use a decrement rate of 0.95 every 1710 steps. The node and edge inputs are namely embedded to a size of 64 and 8. All gGRUs and eGRUs have 4 GRU layers with a hidden size of 128. nMLP consists of 2 layers with ReLU activation.

\begin{figure*}[t]
    \centering
    \includegraphics[width=\textwidth]{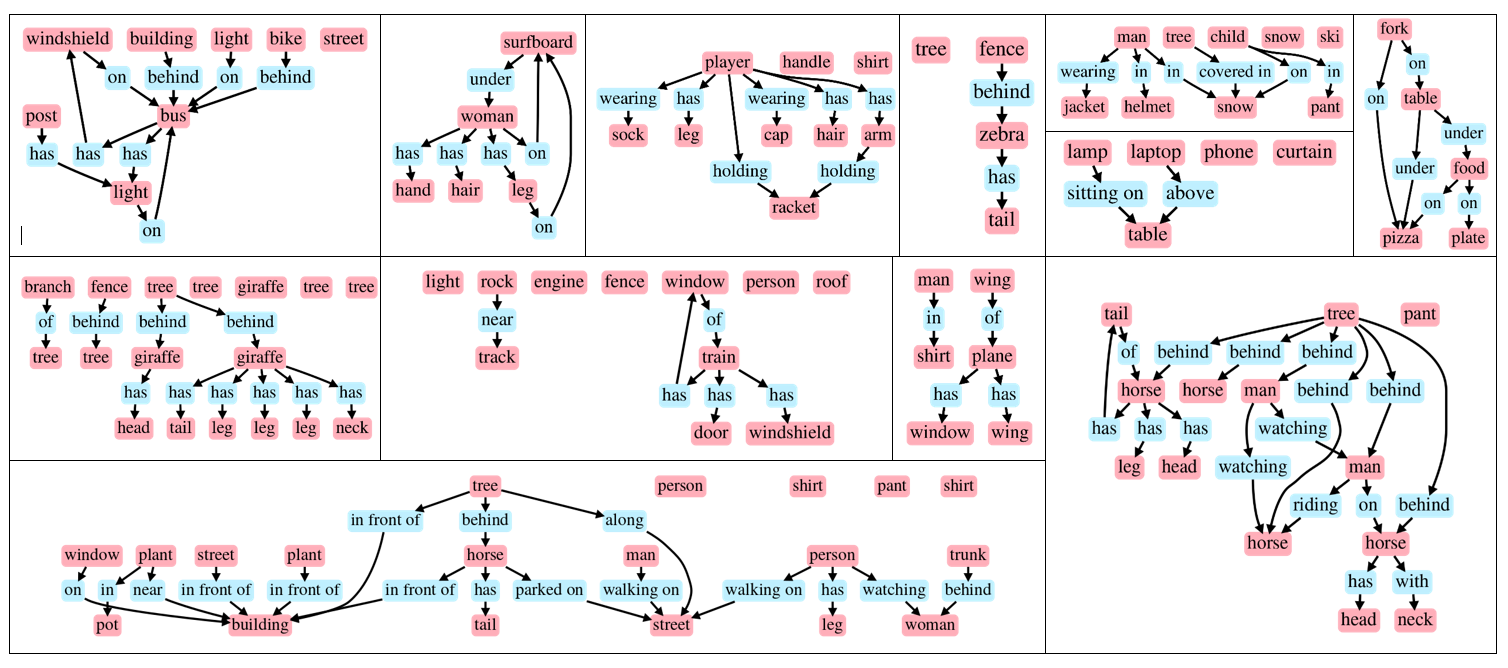}
    \caption{Visualizations of scene graphs generated by SceneGraphGen on Visual Genome. The generated graphs are diverse in size and content and represent reasonable scenes.}
    \label{fig:generated_graphs}
\end{figure*}

\begin{table*}[t]
\centering %
\scalebox{0.8}{
\begin{tabular}{l|c|cc|cccc}
\toprule
\multirow{2}{*} {Model}  & \multirow{2}{*}{Ordering} & \multicolumn{2}{c|}{Graph} &  \multicolumn{4}{c}{Image} \\ 
  &  & MMD node $(\times10^3)$ $\downarrow$  &  MMD graph $(\times10^3)$ $\downarrow$  &  FID $\downarrow$  & IS $\uparrow$ & Precision  $\uparrow$ & Recall $\uparrow$\\ \midrule \midrule
\multirow{2}{*}{GraphRNN \cite{you2018graphrnn}} & BFS &  2.3 & 1.3 & 75.8 & 4.88 & 0.680 & 0.660\\
 & Random &  0.39  & 1.2 & 74.5 & 4.85 & 0.679 & 0.664\\  \midrule
 & BFS &  2.05 & 1.82 & 73.3 & 5.04 & 0.679 & 0.690\\
\nameMethod & Hierarchical & 1.85  & 0.63 & 72.2 & \textbf{5.26} & 0.717 & \textbf{0.714}\\ 
 & Random  &  \textbf{0.37}   &  \textbf{0.11}  & \textbf{71.2}  & 4.95 & \textbf{0.727} & \textbf{0.714} \\ \midrule
Ground Truth &   & 0.018 & 0.023  & 73.0 & 5.22 & 0.693 & 0.707\\ \bottomrule
\end{tabular}}%
\caption{Quantitative evaluation of the graph samples (left) and image samples (right)}
\label{table:osvae_mmd}
\end{table*}

\section{Results}

Here we describe the experiments we carried out to measure the performance of our model. First we introduce our evaluation protocol, including evaluation metrics, baselines and datasets. Then we present qualitative and quantitative results in graph generation. Finally, we illustrate the usefulness of the learned model in three applications: image generation, anomaly detection and graph completion. 

\subsection{Evaluation protocol}
We evaluate our model on the Visual Genome (VG) dataset \cite{krishna2017visual}. Specifically, we use the scene graphs from the widely adopted VG split by Xu \etal~\cite{xu2017scenegraph} containing 150 object categories and 50 relationship categories, and split into train and test sets with 58k and 26k samples respectively. 
Further, since the scene graphs in VG dataset have incomplete relationships, we populate the edges using the Unbiased causal TDE model~\cite{tang2020unbiased} from the SGG benchmark by Kaihua Tang \cite{tang2020sggcode} as it claims less bias in PredCls for rarely occurring relationship categories. Since the different instances of objects (bounding boxes) in scene graphs can correspond to the same underlying object, we use a conservative bounding box intersection-over-union (IoU) of 0.5 along with membership test in hand-made group categories to detect duplicate objects and randomly select one of them. Having duplicate objects also leads to a \textit{child} object (\eg \texttt{shirt}, \texttt{ear}, \texttt{shoe}) having relationships with multiple \textit{parent} objects (\eg \texttt{man}, \texttt{child}). We remove such relationships to a certain extent using heuristics.  
\begin{figure*}
    \centering
    \includegraphics[width=\textwidth]{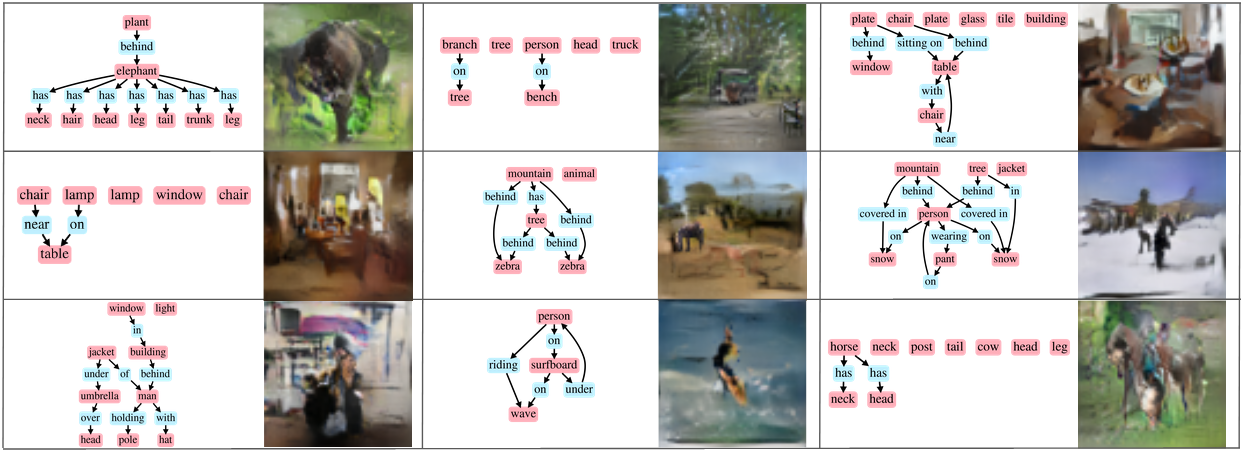}
    \caption{Some examples of 64x64 images synthesized using sg2im on the corresponding scene graphs generated by \nameMethod}
    \label{fig:sg_to_images}
\end{figure*}
\begin{figure*}
    \centering
    \includegraphics[width=0.98\textwidth]{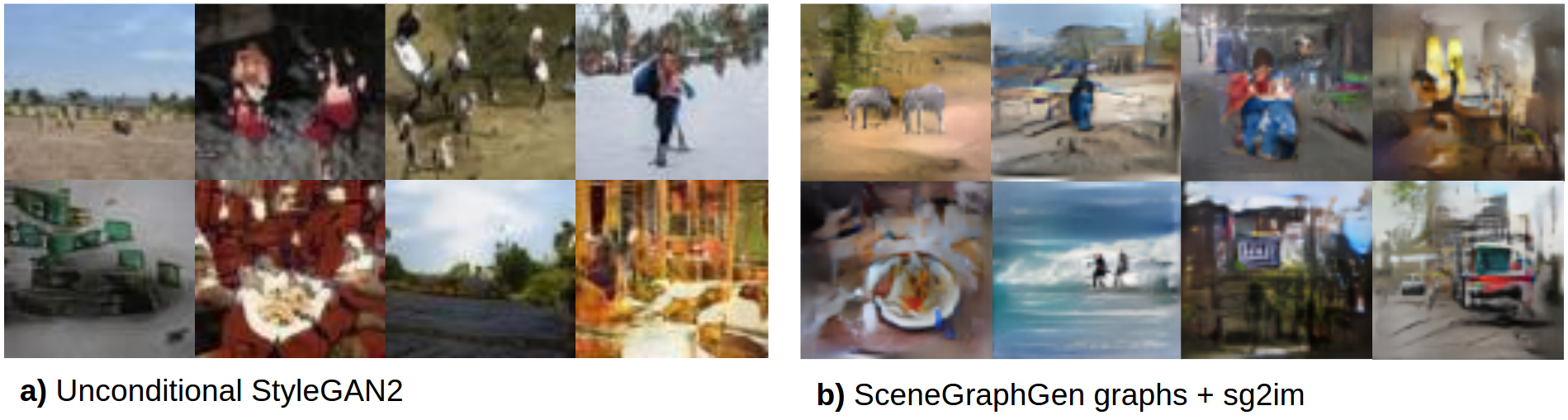}
    \caption{Some examples of 64x64 resolution images synthesized via a.) Unconditional StyleGAN trained on Visual Genome (left) b.) sg2im on scene graphs generated by \nameMethod trained on Visual Genome (right).}
    \label{fig:uncond_vs_cond}
\end{figure*}

Since there are no existing work for unconditional scene graph generation, we compare our model with GraphRNN, adapted to include node and edge categories as well as edge directions. In contrast, SceneGraphGen includes node information during edge generation and conditioning of $E^{from}$ edges on $E^{to}$ edges. We also test our model under various node ordering schemes, namely BFS order, hierarchical order and random order. The BFS node ordering as introduced in GraphRNN~\cite{you2018graphrnn}, visits the graph in a Breadth-First Search order. The hierarchical order sorts the nodes based on their belonging relationships, \ie background (\texttt{sky}), objects/beings (\texttt{elephant}), parts (\texttt{arm}).

Evaluating graph generative models based on sample quality is a difficult task \cite{theis2016note}, as it requires a comparison of the generated set against the test set. We formulate such comparison between the generated and test dataset using Maximum Mean Discrepancy (MMD), which is defined between two distributions $p$ and $q$, and for a given kernel $k$ as
\begin{equation}
\begin{split}
\text{MMD}^2(p, q) = & \mathbb{E}_{x,y\sim p}[k(x,y)] - 2\mathbb{E}_{x\sim p, y \sim q}[k(x,y)] \\& + \mathbb{E}_{x,y\sim q}[k(x,y)].
\end{split}
\end{equation}
\noindent Since MMD has not been used with kernels for directed and labeled graphs, we devise two kernels for comparing any two samples of scene graphs:

\noindent \textbf{Random-walk graph kernel}, inspired from \cite{graphkernel}, computes the similarity between two scene graph samples by comparing the \textit{directed} random walks generated from the graphs.

\noindent \textbf{Object set kernel} computes the similarity between the sets of objects in the scene graphs by comparing the object categories and the number of instances of the objects.

We refer to these metrics as MMD graph and MMD node. MMD graph compares the distributions based on overall graph similarity between samples (triplets, bigger clusters, etc) and MMD node compares the distributions based on the similarity between the respective set of objects (co-occurrence, number of instances, etc). We verify the metrics by observing their outcome on randomly corrupted datasets and the test dataset (Table \ref{fig:mmd_verification}). A dataset corruption of x\% is done by randomly choosing x\% of the nodes and edges in the graph and replacing them with a random label. As expected, MMD values between two disjoint sets of test datasets as well as between two fully-corrupted datasets are very low, as the comparing datasets are very similar. As the level of corruption increases, the MMD metrics between test and corrupted set increases, since the distributions become more dissimilar.
\begin{table}[h!]
\centering
\resizebox{0.9\columnwidth}{!}{
\begin{tabular}{c|c|c}
\toprule
Comparison & MMD node  $\downarrow$ & MMD graph $\downarrow$\\ \midrule
test Vs test &  0.018  &   0.023            \\ 
100\% corrupt Vs 100\% corrupt & 0.11    &    0.0098     \\ 
test Vs 20\% corrupt &  6.0   &   3.7       \\ 
test Vs 50\% corrupt &  10   &   6.3       \\ 
test Vs 100\% corrupt & 44    &  25        \\ \bottomrule
\end{tabular}}
\caption{Sanity checks for the MMD metrics ($\times10^3$), comparing a ground truth set of graphs (test) against a randomly corrupted set.}
\label{fig:mmd_verification}
\end{table}

We further evaluate SceneGraphGen by assessing the quality of images generated by the generated scene graphs. Specifically, we generate $64\times64$ images using the sg2im \cite{johnson2018image} model with the scene graphs generated by SceneGraphGen as input. For assessing the sample quality, we use established metrics in image generation quality, like the Frechet Inception Distance (FID) \cite{Heusel2017GANsTB}, Inception Score (IS) \cite{Salimans2016ImprovedTF}, Precision ($F_{1/8}$) and Recall ($F_{8}$) scores \cite{sajjadi2018assessing}. We compare the generated images against the ground truth Visual Genome images.

\begin{figure*}[t]
\begin{minipage}{0.36\textwidth}
    \includegraphics[width=\linewidth]{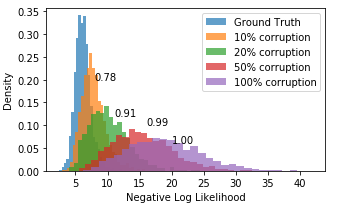}
    \end{minipage}
    \begin{minipage}{0.6\textwidth}
    \includegraphics[width=\linewidth]{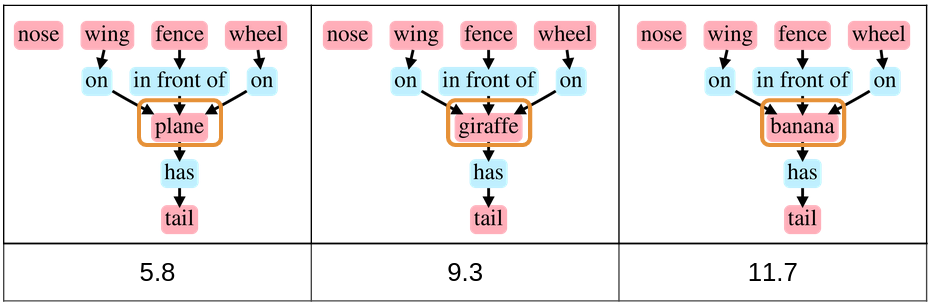}
    \end{minipage}
    \caption{\textbf{Anomaly detection using SceneGraphGen} \emph{Left:} The distribution of negative log-likelihood (NLL) of datasets under varied levels of corruption. The value next to each distribution is the area under ROC curve. \emph{Right:} An example with increasing value of NLL.}
    \label{fig:anomaly}
\end{figure*}  
\begin{figure*}
    \centering
    \includegraphics[width=\textwidth]{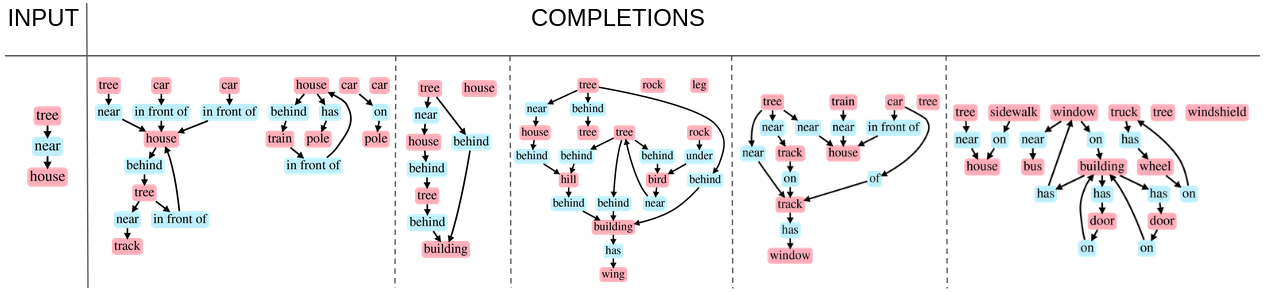}
    \caption{Samples of diverse completed scene graphs from an input partial scene graph.}
    \label{fig:completion}
\end{figure*}

\subsection{Graph generation}
SceneGraphGen can generate realistic and diverse samples of scene graphs. Figure \ref{fig:generated_graphs} shows some visualizations of the generated graphs. We observe that the model can generate graphs corresponding to both indoor (\texttt{table}, \texttt{food}) and outdoor scenes (\texttt{building}, \texttt{fence}, \texttt{horse}), while generating reasonable object co-occurrences and semantically meaningful relationships between the objects. SceneGraphGen outperforms GraphRNN both in terms of MMD graph and MMD node (Table \ref{table:osvae_mmd}, left), showing that node information significantly improve edge generation. We also observe that random ordering outperforms other ordering schemes for scene graph data. Note that BFS and hierarchical ordering are biased towards some node orderings due to their dependence on node connectivity and node semantics respectively. Such bias is not useful for a dataset like Visual Genome which varies a lot in structure and little regularity assumptions can be made. We believe that a random ordering performs better as it has no such bias. We refer the readers to the supplement for a comparison of data statistics of generated dataset and test dataset.

\subsection{Image generation from generated graphs}

Here we want to show that the graphs generated from \nameMethod can be used to synthesize novel images. We use sg2im \cite{johnson2018image}, an off-the-shelf method for image generation from scene graphs to translate the learned graphs in the image domain. We dub this method sg2im-SGG. Figure \ref{fig:sg_to_images} shows some examples of novel scenes generated by the corresponding newly generated scene graphs. Quantitative evaluation on the generated images (Table~\ref{table:osvae_mmd}, right) shows that SceneGraphGen outperforms GraphRNN on all metrics, and is fairly similar to ground truth, \ie images generated from the ground truth test set.

We additionally compare these images against a state-of-the-art model in unconditional image generation \cite{karras2020analyzing}. For fairness, we train the StyleGAN2 model on Visual Genome on the same training split used to train sg2im. Figure \ref{fig:uncond_vs_cond} illustrates results for both methods. We observe that, though StyleGAN2 generates images in good quality for simple image constellations (\eg landscapes), it fails to capture semantic properties in more complex scenes (\eg multiple instances of green screens or indistinguishable objects). On the contrary the images produced by the generated graphs have more well grounded compositions, as dictated by the scene graphs. We also compare the object statistics of StyleGAN vs sg2im-SGG by detecting objects using off-the-shelf Faster-RCNN model trained on COCO. The detector was able to detect a total of 50 object categories for sg2im-SGG images as compared to 40 categories for StyleGAN images, illustrating that using scene graphs as an intermediate representation aids in generating more semantically diverse scenes. Additionally, comparing the object occurrences against the ground truth test set of VG, shows that sg2im-SGG is more in line with the ground truth compared to the StyleGAN2 model, with the ground-truth average error of 1.2 as compared to 1.4. We refer the readers to supplement for a comparison of object occurrence bar plots.

\subsection{Anomaly detection}
\label{section:anomaly}
We demonstrate that the learned model can also be used to detect anomalies in scenes. We evaluate the likelihood over the test dataset using SceneGraphGen, and detect anomalies as scene graphs whose likelihoods are outliers to the likelihood distribution. We illustrate the effectiveness of likelihood in anomaly detection by computing the negative log-likelihood (NLL) over datasets with various levels of node and edge corruption. We compute the area under ROC curve (AUROC) by using NLL scores to classify corrupt samples as anomalies, as a measure of effectiveness of NLL in anomaly detection~\cite{hendrycks2019deep}. Figure \ref{fig:anomaly} (left) shows the NLL distribution along with AUROC values. The respective GraphRNN plot can be found in the supplement. Figure \ref{fig:anomaly} (right) illustrates an example of anomaly, where the more surprising samples lead to a higher NLL value.

\subsection{Scene graph completion}
SceneGraphGen can also generate a scene graph conditioned on a partial graph input, which allows for a more guided generation of scenes. Starting with a partial graph, we randomly sample a sequence of nodes/edges ($X_i$) conditioned on the previous sequence ($X_{i-1}$) -- same as for graph generation --
which accumulates stochasticity at each step leading to diverse completions. Figure \ref{fig:completion} shows how the model is capable of generating diverse possible scene graph completions starting from the same triplet \texttt{tree} - \texttt{near} - \texttt{house}. 
For more examples, please refer to the supplement.%
\section{Conclusions}
\label{sec:conclusion}

We introduced SceneGraphGen, a model for unconditional scene graph generation. We showed that the model is capable of learning semantic patterns from real scenes, and synthesizing reasonable and diverse scene graphs. The model outperforms the baselines on graph-based metrics as well as image-based metrics. We demonstrated the applicability of the learned model and the generated graphs in image generation, detection of out-of-distribution samples and scene graph completion. Future work will be dedicated to exploring conditional variants of the model, such as using text descriptions to constrain the scene category. 

\section{Acknowledgement} We gratefully acknowledge the
Deutsche Forschungsgemeinschaft (DFG) for supporting
this research work, under the project 381855581. We are also thankful to Munich Center for Machine Learning
(MCML) with funding from the Bundesministerium für Bildung und Forschung (BMBF) under the project 01IS18036B.

\newpage

{\small
\bibliographystyle{ieee_fullname}
\bibliography{egbib}
}

\newpage 
\clearpage
\setcounter{section}{0}
\renewcommand*{\thesection}{\Alph{section}}

\section{Supplementary Material}

\noindent This document supplements our paper \emph{Unconditional Scene Graph Generation} with dataset-level statistics, the mathematical description of the MMD kernels, and additional results on the different applications.

\subsection{Dataset-level statistics}

\begin{figure*}[htp]
\centering
    \subfloat[Object occurrence]{
    \begin{tabular}[b]{c}
    \includegraphics[width=0.9\linewidth]{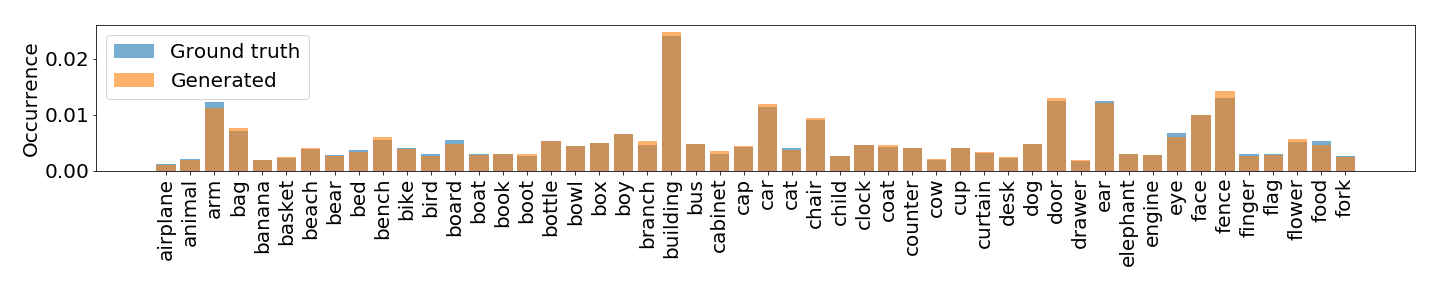} \vspace{-4mm}\\
   \includegraphics[width=0.9\linewidth]{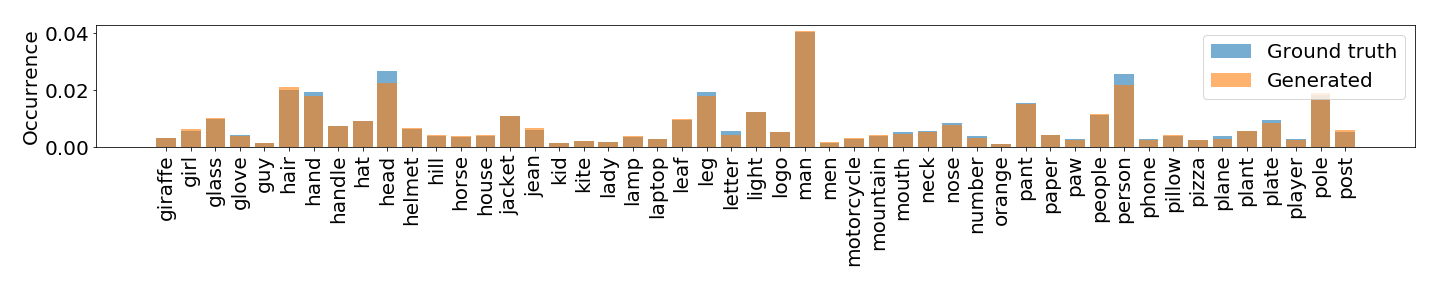} \vspace{-4mm}\\
   \includegraphics[width=0.9\linewidth]{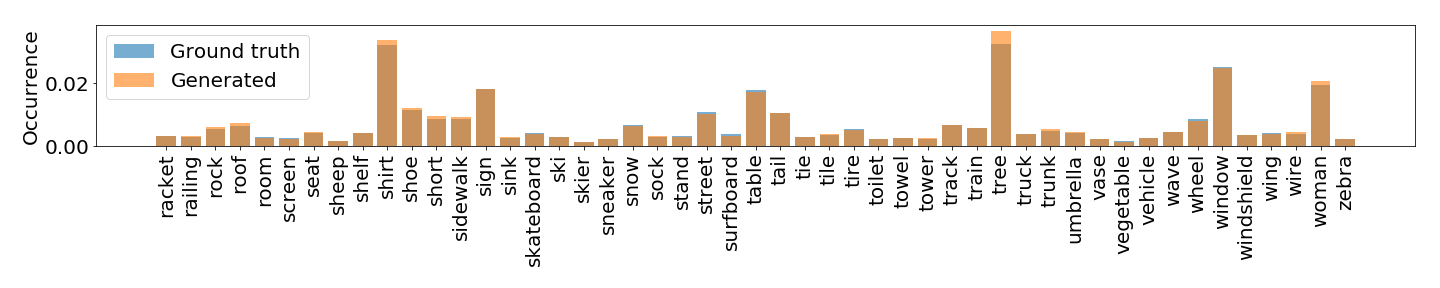}
   \end{tabular}} \\
   \subfloat[Relationship occurrence]{ \includegraphics[width=0.9\linewidth]{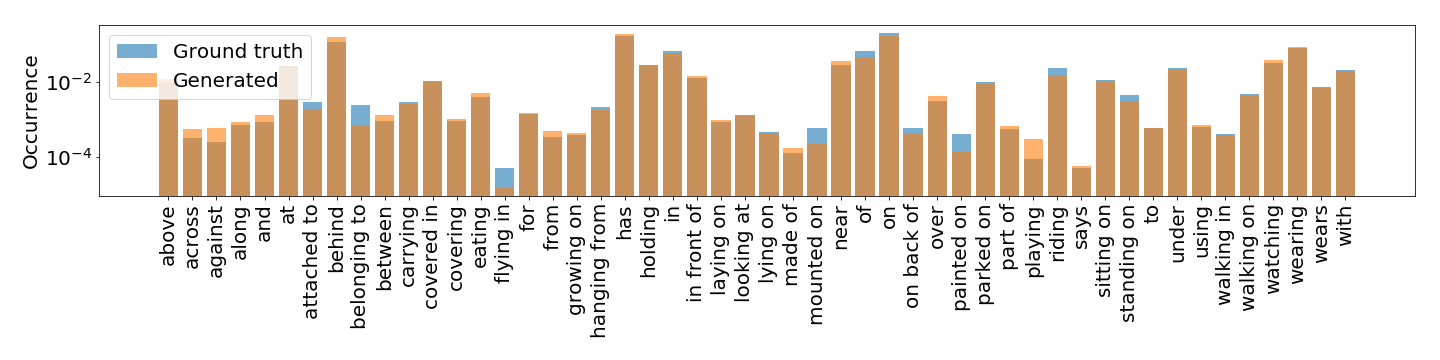}} \\
    \subfloat[Object Co-Occurrence]{\includegraphics[width=0.8\linewidth]{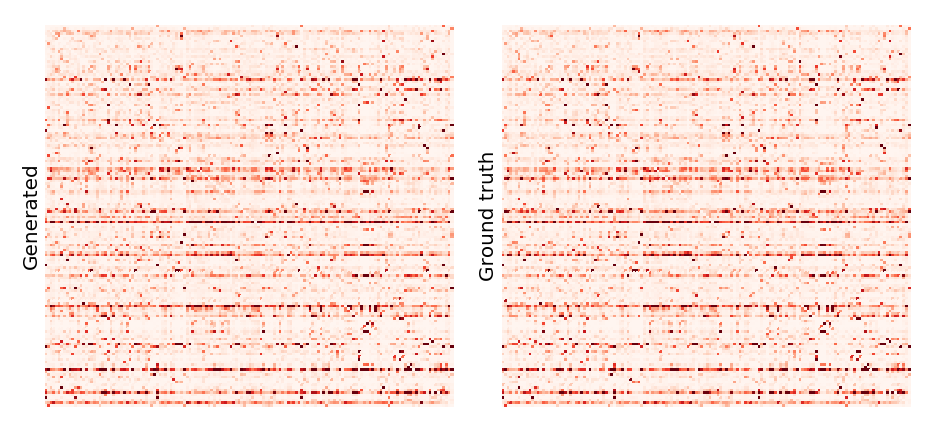}}
\caption{Comparison of the dataset-level statistics of generated scene graphs against ground truth scene graphs from Visual Genome. a.) Object Occurrence, b.) Relationship Occurrence, c.) Object Co-Occurrence}
\label{fig:statistics}
\end{figure*}

\begin{figure*}[htp]
\centering
\begin{tabular}[b]{c}
    \includegraphics[width=0.9\linewidth]{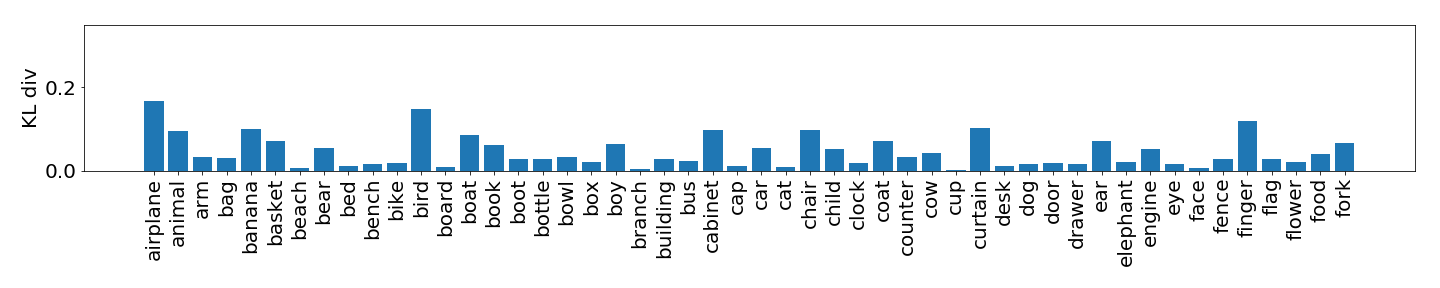} \vspace{-4mm}\\
   \includegraphics[width=0.9\linewidth]{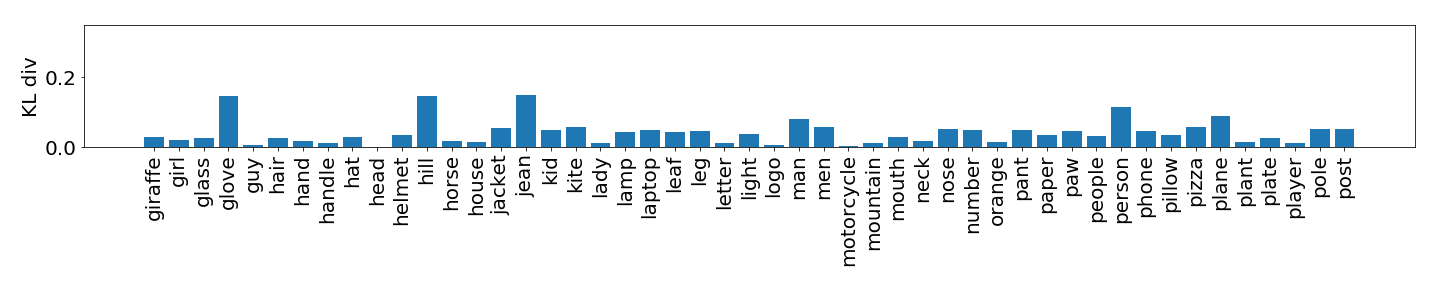} \vspace{-4mm}\\
   \includegraphics[width=0.9\linewidth]{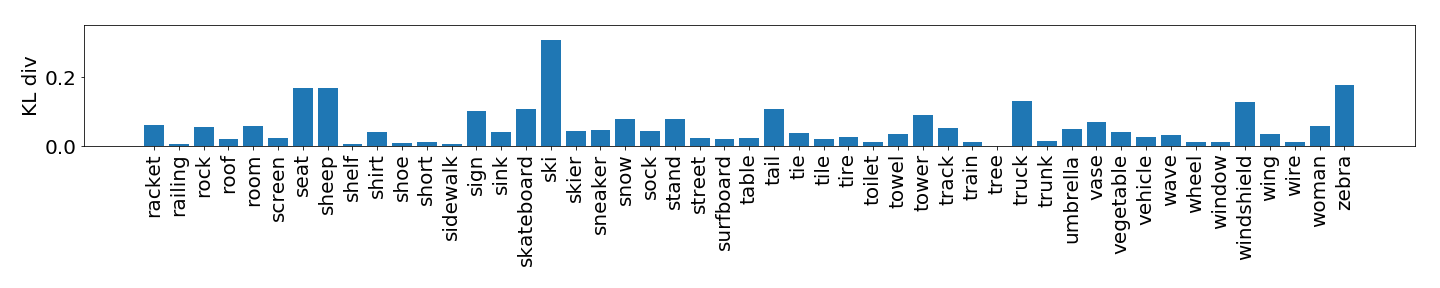}
\end{tabular}
\caption{KL divergence of the generated dataset from test dataset, which compares the object count distribution (number of instances of a particular object category per scene) for each object category}
\label{fig:statistics1}
\end{figure*}

In addition to the MMD metrics (sample level comparison), we present the statistics to compare the generated and test samples on the dataset level. We compare 20k samples of generated scene graphs from SceneGraphGen, against the ground truth \ie the test dataset from Visual Genome. Figure \ref{fig:statistics} reports different patterns such as object occurrence (a), relationship occurrence (b) and object co-occurrence (c). Object occurrence computes the occurrence probability of each object label over the whole dataset. Similarly, relationship occurrence computes the occurrence probability of each relationship label over the whole dataset. The object co-occurrence is computed by collecting the frequency (normalized) with which two \textit{different} object labels co-occur in the same scene graph (scene). For improved visibility of co-occurrence patterns, we use a maximum threshold of 0.05. In all measurements in Figure~\ref{fig:statistics}, we observe similar patterns between the generated and ground truth datasets. Since each object category can occur multiple times in an image (instances), we compare the count distribution (1, 2, so on) for each object category using Kullback-Leibler (KL) divergence of generated dataset from the test dataset. Figure \ref{fig:statistics1} shows the KL divergence of object count distribution for each object category, with a low average value of 0.048.

\subsection{Object occurrences in the generated images}
\begin{figure*}[htp]
\centering
\includegraphics[width=0.9\linewidth]{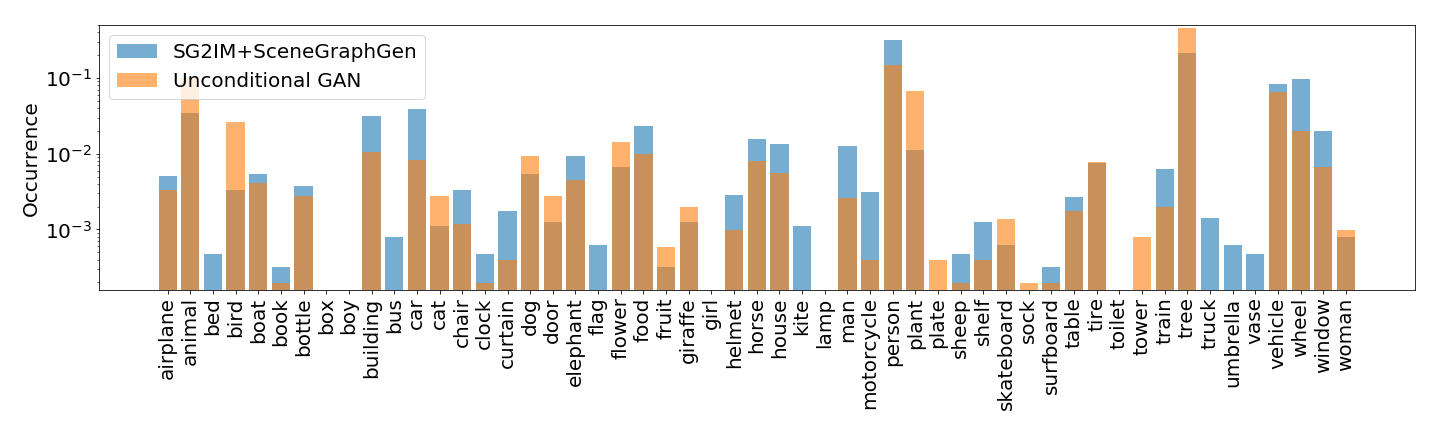}
\caption{Comparison of occurrence of objects detected by Faster R-CNN on the images generated by Unconditional-GAN (StyleGAN2) vs. sg2im+SceneGraphGen model}
\label{fig:detected_stats}
\end{figure*}

Figure \ref{fig:detected_stats} shows the object occurrence of detected objects (using FasterRCNN) in the images generated by StyleGAN (unconditional) vs the images generated by sg2im on scene graphs generated by SceneGraphGen (sg2im-SGG). We observe that sg2im-SGG generates images with more objects detected and better object statistics than StyleGAN. The FID of StyleGAN on Visual Genome is however $66.3$, so better than sg2im-SGG, which we attribute to the respective image generative models, instead of the quality of input scene graph.

\subsection{Additional examples from applications}
\begin{figure*}[htp]
\centering
\includegraphics[width=\linewidth]{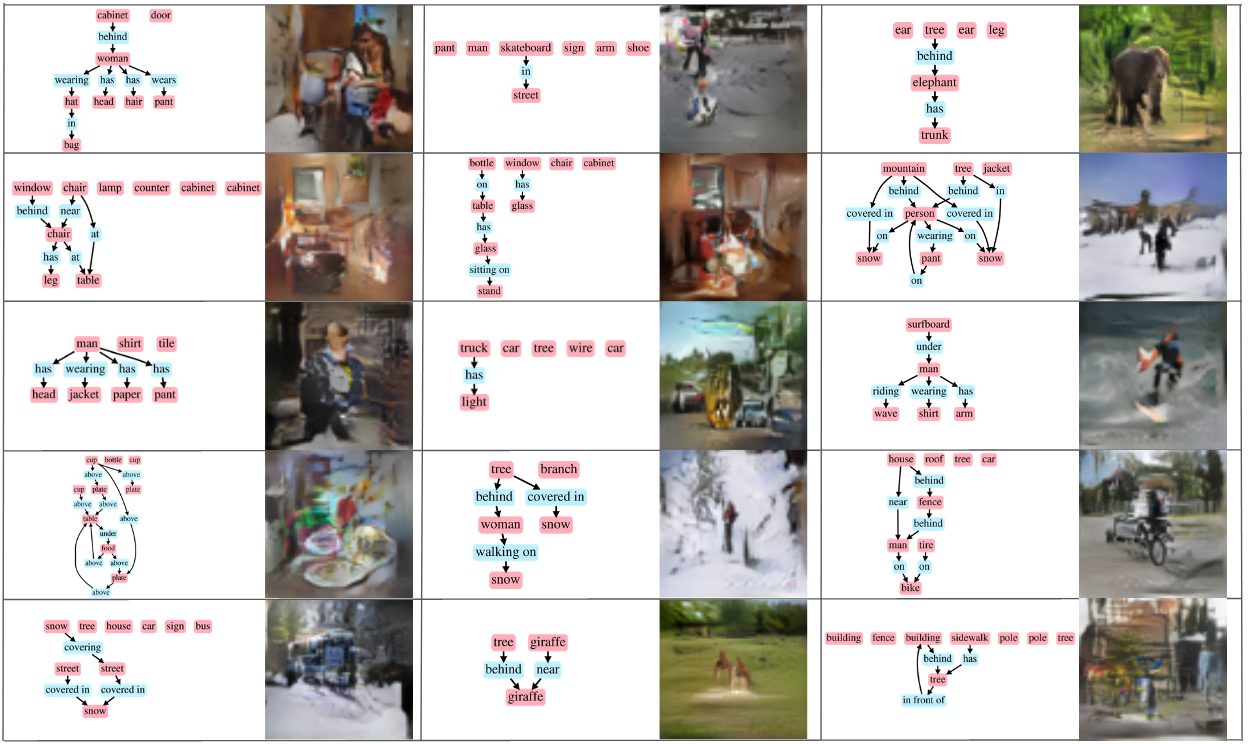}
\caption{Additional examples of 64$\times$64 images generated using sg2im on the corresponding scene graphs generated by SceneGraphGen}
\label{fig:sg_to_images_supp}
\end{figure*}

\begin{figure*}[htp]
\centering
\includegraphics[width=\linewidth]{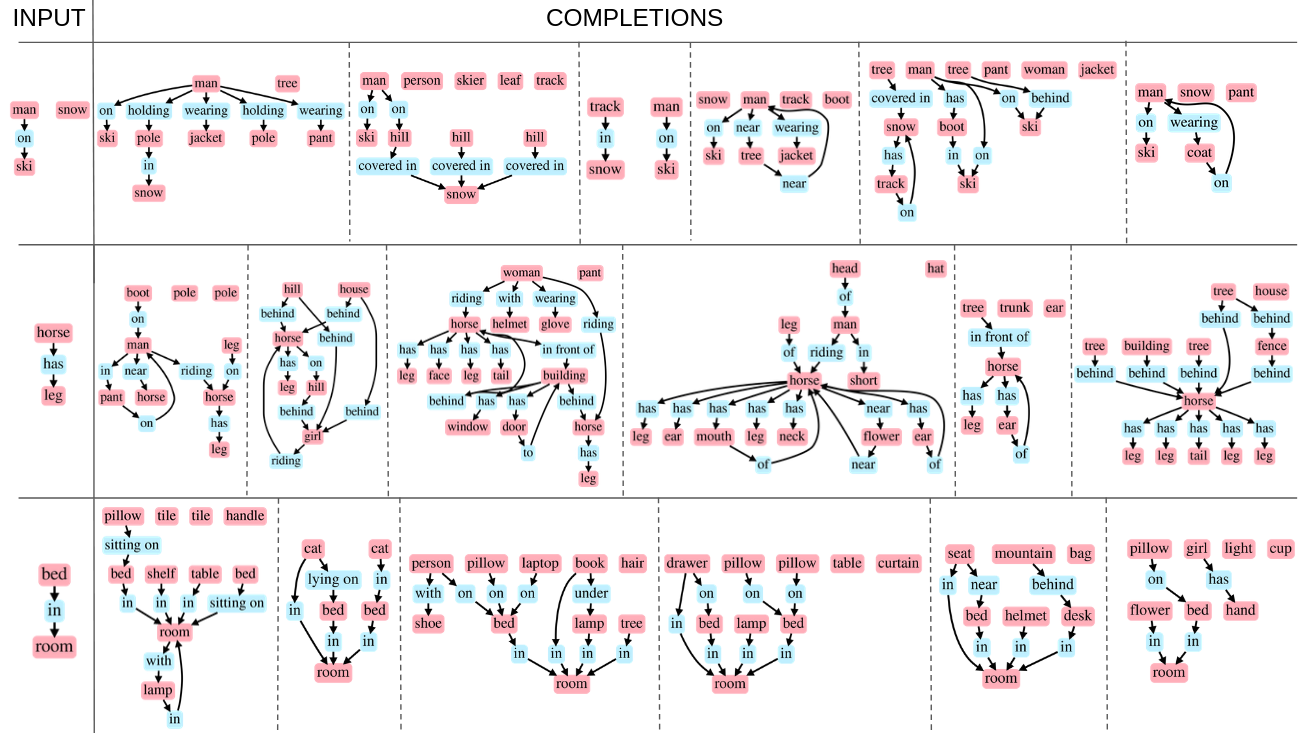}
\caption{Additional examples of scene graph completions from an initial partial scene graph using SceneGraphGen}
\label{fig:completion_supp}
\end{figure*}

We provide some additional examples for image generation in Figure \ref{fig:sg_to_images_supp}, and scene graph completion in Figure \ref{fig:completion_supp}.

\subsection{Mathematical formulation of the MMD kernels}

\noindent Here we give the details on the kernels used for the MMD evaluation. 

\subsubsection{Random walk graph kernel}
A general formulation for random walk kernel is adopted from \cite{graphkernel}, which allows freedom to choose suitable node and edge kernels. In two graphs $G_a$ and $G_b$, we want to compare two nodes $r$ and $s$ respectively. We can compare these two nodes by comparing all walks of length $p$ in $G_a$ starting from $r$ against all walks of length $p$ in $G_b$ starting from $s$. The similarity between each walk-pair can be performed by comparing the respective nodes and edges encountered in the walks using suitable kernels. The kernel to compare any two nodes is given by
\begin{equation}
\begin{split}
\label{equation:graphkernel}
    k^p_{R}(G_a, G_b, r, s) = & {\sum_{
    \begin{subarray}{c} 
    (r_1, e_1, .., e_{p-1}, r_p)\in W^p_{G_a}(r) \\
    (s_1, f_1, .., s_{p-1}, f_p)\in W^p_{G_b}(s) 
    \end{subarray}}
    \Big[ k_{node}(r_p, s_p)} \\
    & \prod_{i=1}^{p-1} k_{node}(r_i, s_i) k_{edge}(e_i, f_i) \Big]\\
    \end{split}
\end{equation}
To compare the overall structure, the kernel in Equation \ref{equation:graphkernel} is summed over all pairs of nodes, and normalized with maximum of the kernel evaluation of each graph and itself.

\begin{equation}
    k^p_G(G_a, G_b) = \sum_{\begin{subarray}{c} 
    r \in V_{G_a} \\
    s \in V_{G_b}
    \end{subarray}}
    k^p_{R}(G_a, G_b, r, s)
\end{equation}
\begin{equation}
    k^{N}_G(G_a, G_b) = \frac{k_G(G_a, G_b)}{\text{max}(k_G(G_a, G_a), k_G(G_b, G_b))}
\end{equation}
For comparing nodes, we use the simple Kronecker delta function which is 1 when the node categories match and 0 otherwise, \ie $k_{node}(r, s) = \delta(r, s)$. However, since there are multiple nodes with the same category, the importance of the nodes in a graph will be lower for the category with one occurrence and higher for multiple occurrences. In fact, the importance of a category with multiple occurrences should diminish as the occurrences increase. For this purpose, the node kernel is normalized with the frequency of occurrence in a graph. The node kernel is given by
\begin{equation}
\begin{split}
    k^N_{node}(r, s) = \sigma(r)\sigma(s) k_{node}(r, s),\\
    \text{where} \quad \sigma(s) = \frac{1}{\sum_{s \in V_{G_s}}k_{node}(r, s)}\\
\end{split}
\end{equation}
For comparing edges, we use the Kronecker delta function, \ie $k_{edge}(p, q) = \delta(p, q)$.

\subsubsection{Object set kernel}
We want to compare two \textit{sets} of object instances. Hein \etal~\cite{setkernels} showed that for a domain set $\mathcal{X}$, two sets $A \in \mathcal{X}$, $B \in \mathcal{X}$, a positive definite kernels $k_{label}$ and $k_{count}$, we can define a general set kernel between $A$ and $B$ as:
\begin{equation}
    k_{set}(A, B) = \sum_{x \in \mathcal{X}} \sum_{y \in \mathcal{X}} k_{label}(x, y) k_{count}(A(x), B(y))
\end{equation}
We choose $k_{label}(x, y) = \delta(x, y)$ as the Kronecker delta function which is one when both $x$ and $y$ have the same object categories. $A(x)$ is the number of times element $x$ appears in $A$ and $B(y)$ is the number of times element $y$ appears in $B$. $k_{count}$ is defined as
\begin{equation}
    k_{count}(A(x), B(y)) = \frac{1}{1+ |A(x)-B(y)|}
\end{equation}
$k_{count}$ is the generalized t-student kernel \cite{tstudentkernel,tstudent2}. This formulation allows us to capture when the two sets have the same object category member as well as how similar are the counts of those object category members. Similar to the graph kernel above, the object-set kernel is also normalized with the maximum of kernel evaluation of each object set with itself.
\begin{equation}
    k^{N}_{set}(A, B) = \frac{k_{set}(A, B)}{\text{max}(k_{set}(A, A), k_{set}(B, B))}
\end{equation}

\subsection{Anomaly detection comparison}

\begin{figure}[h!]
\begin{center}
   \includegraphics[width=\linewidth]{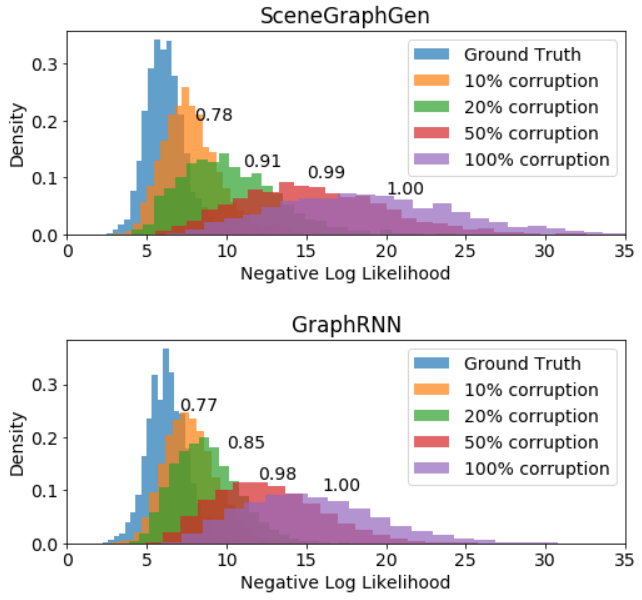}
\end{center}
   \caption{Distribution of NLL under varied levels of corruption}
\label{fig:anomaly_comparison}
\end{figure}

Figure \ref{fig:anomaly_comparison} demonstrates a comparison between SceneGraphGen and the GraphRNN baseline on the NLL plot, extending Figure 6 (left) in the main paper. Our interpretation is that our model is more sensitive to the level of dataset corruption compared to the baseline, \ie the NLL gap between the different levels is larger than for the baseline.

\subsection{Checking for overfitting}
\begin{figure}[h!]
\begin{center}
   \includegraphics[width=\linewidth]{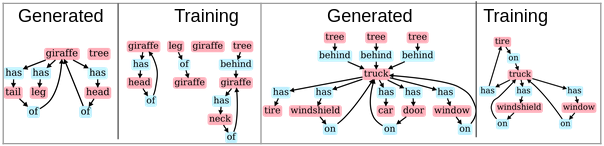}
\end{center}
\caption{Examples of generated graphs as well as closest sample from the training set.}
\label{fig:overfit}
\end{figure}

Figure \ref{fig:overfit} shows examples of generated graphs and the respective nearest graph (via graph kernel comparison) from training data. The graphs are not identical, \ie the model is not reproducing examples from the training set. 
  
\end{document}